*Research Article*

# Development of Comprehensive Devnagari Numeral and Character Database for Offline Handwritten Character Recognition

## Vikas J. Dongre and Vijay H. Mankar

*Department of Electronics & Telecommunication Engineering, Government Polytechnic, Nagpur 440 001, India*

Correspondence should be addressed to Vikas J. Dongre, dongrevj@yahoo.co.in

In handwritten character recognition, benchmark database plays an important role in evaluating the performance of various algorithms and the results obtained by various researchers. In Devnagari script, there is lack of such official benchmark. This paper focuses on the generation of offline benchmark database for Devnagari handwritten numerals and characters. The present work generated 5137 and 20305 isolated samples for numeral and character database, respectively, from 750 writers of all ages, sex, education, and profession. The offline sample images are stored in TIFF image format as it occupies less memory. Also, the data is presented in binary level so that memory requirement is further reduced. It will facilitate research on handwriting recognition of Devnagari script through free access to the researchers.

## 1. Introduction

With the advent of development in computational power, machine simulation of human reading has become a topic of serious research. Optical character recognition (OCR) and document processing have become the need of time with the popularization of desktop publishing and usage of internet. OCR involves recognition of characters from digitized images of optically scanned document pages. The characters thus recognized from document pages are coded with American Standard Code for Information Interchange (ASCII) or some other standard codes like UNICODE for storing in a file, which can further be edited like any other file created with some word processing software. A lot of research has been done in developed countries for English, European, and Chinese languages. But there is a dearth of need to carry out research in Indian languages. One common problem with the research is the need of benchmark database. To facilitate results on uniform data set, several document processing research groups have collected large numeral and character databases to make it available to the fellow researchers around the world. However, such existing databases are available only in few languages such as English,

Japanese, and Chinese [1]. These standard databases include MNIST, CEDAR [2], and CENPARMI in English. Some work is also done for Indic scripts such as Bangla [3], Kannada [4], and Devnagari [5–8]. India is a multilingual and multiscript country having more than 1.2 billion population with 22 constitutional languages and 10 different scripts. Devnagari is the most popular script in India. Hindi, the national language of India which is spoken by more than 500 million population worldwide, is written in the Devnagari script. Moreover, Hindi is the third most popular language in the world [9]. Devnagari is also used for writing Marathi, Sanskrit, Konkani, and Nepali languages.

In a developing country and emerging superpower like India, there is a need for the research and development of its own language technologies. The Department of Information Technology, Government of India, started a program on technology development for Indian languages [10] where language aspects are studied and developed. Another government undertaking Centre for Development of Advanced Computing [11] is actively involved in development of Indian languages fonts, translators. As a result of such initiatives, various research works for automatic recognition of printed/handwritten characters of various Indic scripts are in progress. Some pioneering works on printed Indian scripts include [4, 12] for Bangla, [13] for Kannada, and [14] for Devnagari optical character recognition systems. There exist few studies on handwritten characters of some Indian scripts which include [15–20] for Devnagari characters. Research reviews on Devnagari character recognition are also available which includes [9, 21, 22].

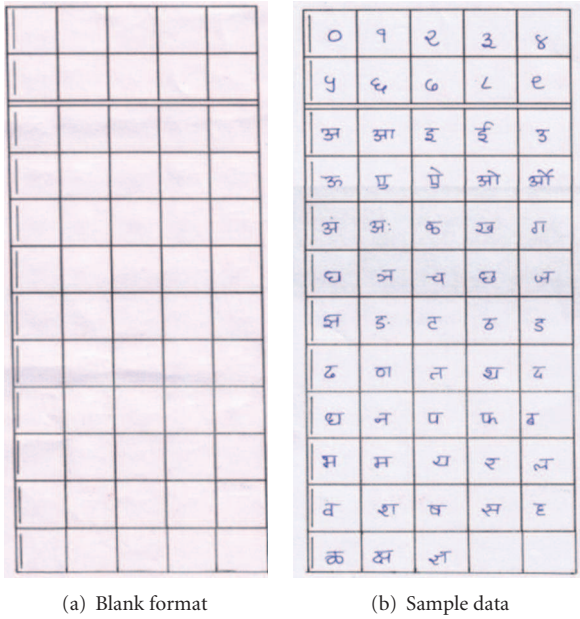

(a) Blank format        (b) Sample data

Figure 1: Handwritten data sample sheet.

Studies are reported on the basis of different databases collected either in laboratory environment or from smaller groups of the concerned population. The effective research work on handwriting recognition for Indic scripts is seriously hampered because of the unavailability of standard/benchmark databases, and those may be used for testing of algorithms and for comparison of results [3].

This paper describes an attempt for generation of a comprehensive database for handwritten Devnagari numerals and characters. This database has been developed with the view to make it available freely to the researcher community as a benchmark database for handwriting recognition research. The printed form of Devnagari numerals, vowels, and consonants are shown in Figures 4, 5, and 6. Sample handwritten form containing numerals and characters collected from a writer is shown in Figure 1. The present paper is organized as follows. Section 2 describes the details of offline database generation. Section 3 discusses statistical analysis of this work. Conclusion and further work direction are discussed in Section 4.

## 2. Devnagari Offline Database Generation Details

*2.1. Data Collection.* A sample A4 size sheet having blank boxes was designed. Persons of various ages, sex, education, and occupation were requested to write Devnagari numbers and characters. The only imposed restriction was that the character or numeral stroke should not touch the boundary of the boxes on the sheet and the vertical line made in the first box of every row. No restriction was imposed regarding colour of ink, thickness of lines, sequence of characters, and type of pen like ball pen or ink/gel pen. In case pen was not available with the writer, it was supplied at random from a set of different types of pens. The data was collected from 750 writers which included students of schools and colleges, office staff, workers, housewives, and senior citizens. The writers were carefully chosen to make the database representative. Persons of various languages and educational background like Marathi, Hindi were involved for writing on the blank sheets. Data was also collected from persons waiting in railway reservation centers and hospitals, which consists the mixture of all the categories mentioned earlier. Option of disclosing personal information was left to the writers so as to keep them free from stress that writing must be legitimate. Figure 1 shows sample handwritten data written by a writer.

*2.2. Data Preparation.* The A4 size paper sheet having the data written by various writers (Figure 1(b)) is digitized using Canon Canoscan Lide 100 flatbed scanner at 300 dpi. The images were stored in JPG format. It is cumbersome and time-consuming task to separate isolated symbols from the scanned image. Hence, various software modules were developed in Matlab to perform this task. The overall procedure is explained in the following. In all of the 750 samples, sheets are used in this work. Scanned images of the original paper sheets are also preserved in the original form for future use.

(1) Gray scale image is converted to binary for simplicity. In pattern recognition, we are concerned with shape and size of the object and not the color or gray level details. This also reduces data storage requirements as well as computation time.

(2) Isolated pixels (noise) are removed.

(3) The boundaries around the numerals and characters are removed using simple logic that it is the first and biggest continuous object. Other isolated groups of pixel are considered as desired data.

(4) Various rows are segmented using horizontal histogram approach [23]. Zero pixels in the histogram indicate separation of various rows. Each row is separately processed. Each row begins with vertical line as first object, which is ignored. This is used specially for preserving the dot present as a part of character अँ in Devnagari script. Otherwise, this character resembles with अ and all the images of अँ are lost.

(5) Useful characters segmented are stored in individual files. TIFF format is used for this purpose.

(6) The separated symbols are visually checked for proper shapes before sorting and storing in proper folders. 60 folders are formed for storing 10 numeral

(a) Valid numeral database

(b) Valid character database

Figure 2: Valid sample database.

(a) Invalid isolated strokes

(b) Ambiguous characters

Figure 3: Invalid discarded data.

databases and 50 character databases. A few samples of isolated numerals and characters from the present database are shown in Figure 2.

(7) Various image symbol files are serially numbered for further convenient use. Figure 7 shows size of numeral database for each numeral, and Figure 8 shows size of database for each character.

## 3. Statistics of Data Generated

Some Devnagari compound characters are not widely used in modern writing (e.g., ड़ and ऋ). Some characters are written in more than one way, for example, १ as ?, ८ as ८, and ९ as ६. The database mostly contains first form of the numeral as it is written by most of the writers. Second form of the character is also written by few writers which is preserved in the database. The researcher may separate such data as per his/her need.

The ideal Devnagari script consists of curves and connected lines. Lines are not isolated from main symbol. But in practice, the handwritten documents and the number of strokes are unintentionally isolated due to inaccurate writing of writers. This imposes serious problems in document segmentation and further recognition. In the character segmentation stage, isolated strokes of modifiers are mistakenly considered as individual symbol and thus stored separately. Correctly segmented numerals and characters are shown in

Figure 4: Devnagari Numerals.

Figure 5: Devnagari Vowels and modifiers.

Figure 6: Devnagari Consonants.

Figure 2. Isolated strokes and symbols in the handwritten document are shown in Figure 3(a). These captured strokes are rejected after visual inspection and removed from database. Also, ambiguous numerals or characters which may belong to more than one category are removed from database. Figure 3(b) shows such possible characters. Various characters are containing open curves and lines. Such characters cannot be uniquely categorized. Hence, they are also rejected. Some characters are improperly written by writers.

| Data base no. | Symbol | Frequency | Data base no. | Symbol | Frequency |
|---|---|---|---|---|---|
| 0 | ० | 490 | 5 | ५ | 482 |
| 1 | १ | 492 | 6 | ६ | 522 |
| 2 | २ | 634 | 7 | ७ | 458 |
| 3 | ३ | 582 | 8 | ८ | 482 |
| 4 | ४ | 501 | 9 | ९ | 494 |
| | | | | Total | 5137 |

FIGURE 7: Devnagari Numeral database.

Such characters are also rejected In all of the 750 samples, sheets containing all the symbols were processed. Due to the reasons mentioned in the previous paragraph, various databases differ in frequency as shown in Figures 7 and 8.

It can be easily seen that the symbols having the combination of open curve and line (e.g., अं, अः, ङ, ब, त, ज, न, ठ, ण, थ, ध, श, प, ह, क्ष, and ज्ञ) have more chances of ambiguity and incorrectness. Such wrong strokes and ambiguous characters are removed from final database. It may be noted that recognition efficiency for the previously mentioned characters may be poor.

Some characters got wrongly segmented as another valid character due to limitation of segmentation algorithm, for example, र as २, अं as अ, and अः as अ. Hence, it can be observed from Figure 7 that the frequency of numeral २ is more than that of other numerals. On the contrary, frequency of character र is reduced (see Figure 8). It can also be observed from Figure 8 that the frequency of अ is more (878) than that of any other character whereas frequency of अं and अः are much less (195 and 92, resp.). It may be noteworthy that the frequency for अ is even more than that of actual datasets scanned (750 images).

Some characters like ऋ, ङ, ञ, and त्र are rarely used in modern writing. Hence, many writers skipped writing these characters in the blank datasheet provided. So, the frequency for previous characters is very low.

The character ळ is not a part of Devnagari database, rather it is a part of Marathi language which uses Devnagari script. The database for this character is also developed so that it may be useful for research on recognition of Marathi language.

Thus the quantity of numerals and characters in each category of database is reduced and varies as seen from Figures 7 and 8. It can also be observed that the symbol rejection rate is low for numerals than for characters. Hence, numeral recognition efficiency will be much better than character recognition efficiency.

## 4. Conclusion and Future Work

In this paper, we have generated a comprehensive database for Devnagari numerals and characters. Database of 5137 symbols is generated for numerals, and database of 20305 symbols is generated for characters. It is found that some symbols obtained need to be rejected as the writings of many persons are not recognizable by visual inspection. It will be impossible for computer software to recognize such symbols. The data images are stored in binary level and TIFF format for efficient storage and computational needs. This database will be further grown with more samples from variety of writers. Also, the database will be categorized as training set and test set randomly in near future. This database will be made freely available on http://code.google.com/p/devnagari-database/. This will surely help the research community for benchmarking their research results.

| Data base no. | Symbol | Frequency | Data base no. | Symbol | Frequency |
|---|---|---|---|---|---|
| 10 | अ | 878 | 35 | ड | 522 |
| 11 | आ | 475 | 36 | ढ | 405 |
| 12 | इ | 496 | 37 | ण | 419 |
| 13 | ई | 484 | 38 | त | 417 |
| 14 | उ | 540 | 39 | थ | 424 |
| 15 | ऊ | 457 | 40 | द | 433 |
| 16 | ऋ | 17 | 41 | ध | 421 |
| 17 | ए | 508 | 42 | न | 437 |
| 18 | ऐ | 474 | 43 | प | 451 |
| 19 | ओ | 447 | 44 | फ | 431 |
| 20 | औ | 439 | 45 | ब | 389 |
| 21 | अं | 195 | 46 | भ | 409 |
| 22 | अः | 92 | 47 | म | 415 |
| 23 | क | 497 | 48 | य | 444 |
| 24 | ख | 458 | 49 | र | 437 |
| 25 | ग | 441 | 50 | ल | 421 |
| 26 | घ | 484 | 51 | व | 467 |
| 27 | ङ | 68 | 52 | श | 344 |
| 28 | च | 429 | 53 | स | 447 |
| 29 | छ | 436 | 54 | ष | 380 |
| 30 | ज | 434 | 55 | ह | 426 |
| 31 | झ | 410 | 56 | ळ | 293 |
| 32 | ञ | 174 | 57 | क्ष | 356 |
| 33 | ट | 470 | 58 | ज्ञ | 390 |
| 34 | ठ | 412 | 59 | त्र | 112 |
| | | | | Total | 20305 |

FIGURE 8: Devnagari Character database.

## Acknowledgments

The authors would like to thank Mrs. Rupali Dongre, Mr. Jitendra Bangari, and Mr. Prashant Kelzare for helping in digitization and sorting of the database. They would also like to thank all the writers who contributed in this database.

## References

[1] T. Saito, H. Yamada, and K. Yamamoto, "On the database ELT9 of hand printed characters in JIS Chinese characters and its analysis," *Transactions of the Institute of Electronics and Communication Engineers of Japan*, vol. J.68-D, no. 4, pp. 757–764, 1985 (Japanese).

[2] J. J. Hull, "A database for handwritten text recognition research," *IEEE Transactions on Pattern Analysis and Machine Intelligence*, vol. 16, no. 5, pp. 550–554, 1994.

[3] B. B. Chaudhuri, "A complete handwritten numeral database of Bangla—a major Indic script," CVPR Unit, Indian Statistical Institute, Kolkata-108, India.

[4] U. Bhattacharya and B. B. Chaudhuri, "Handwritten numeral databases of Indian scripts and multistage recognition of mixed numerals," *IEEE Transactions on Pattern Analysis and Machine Intelligence*, vol. 31, no. 3, pp. 444–457, 2009.

[5] U. Bhattacharya and B. B. Chaudhuri, "Databases for research on recognition of handwritten characters of Indian scripts," in *Proceedings of the 8th International Conference on Document Analysis and Recognition (ICDAR '05)*, pp. 789–793, September 2005.

[6] "Handwritten character databases of Indic scripts," 2012, http://www.isical.ac.in/~ujjwal/download/database.html.

[7] R. Sarkar, N. Das, S. Basu, M. Kundu, M. Nasipuri, and D. K. Basu, "CMATERdb1: a database of unconstrained handwritten Bangla and Bangla-English mixed script document image," *International Journal on Document Analysis and Recognition*, vol. 15, no. 1, pp. 71–83, 2012.

[8] M. P. Kumar, S. R. Kiran, A. Nayani, C. V. Jawahar, and P. J. Narayanan, "Tools for developing OCRs for Indian scripts," in *Proceedings of the Computer Vision and Pattern Recognition Workshop (CVPRW '03)*, pp. 33–38, 2003.

[9] U. Pal and B. B. Chaudhuri, "Indian script character recognition: a survey," *Pattern Recognition*, vol. 37, no. 9, pp. 1887–1899, 2004.

[10] TDIL, 2012, http://www.tdil.mit.gov.in/.

[11] CDAC, 2012, http://www.cdac.in/.

[12] B. B. Chaudhuri and U. Pal, "A complete printed Bangla OCR system," *Pattern Recognition*, vol. 31, no. 5, pp. 531–549, 1998.

[13] A. Aleai, P. Nagbhushan, and U. Pal, "Benchmark Kannada handwritten document database and its segmentation," in *Proceedings of the International Conference on Document Analysis and Research (ICDAR '11)*, pp. 141–145, 2011.

[14] V. Bansal and R. M. K. Sinha, "Integrating knowledge sources in Devanagari text recognition system," *IEEE Transactions on Systems, Man, and Cybernetics Part A*, vol. 30, no. 4, pp. 500–505, 2000.

[15] R. Bajaj, L. Deym, and S. Chaudhuri, "Devnagari numeral recognition by combining decision of multiple connectionist classifiers," *Sadhana*, vol. 27, no. 1, pp. 59–72, 2002.

[16] U. Bhattacharya and B. B. Chaudhuri, "A majority voting scheme for multi-resolution recognition of hand printed numerals," in *Proceedings of the 7th International Conference on Document Analysis and Recognition (ICDAR '2003)*, 2003.

[17] C. V. Jawahar, J. P. Pavan Kumar, and S. S. Ravi Kiran, "A bilingual OCR for Hindi-Telugu documents and its applications," in *Proceedings of the 7th International Conference on Document Analysis and Recognition (ICDAR '2003)*, pp. 1–7, 2003.

[18] R. J. Ramteke, P. D. Borkar, and S. C. Mehrotra, "Recognition of isolated Marathi handwritten numerals: an invariant moments approach," in *Proceedings of the International Conference on Cognition and Recognition*, pp. 482–489, 2005.

[19] T. K. Bhowmik, S. K. Parui, and U. Roy, "Discriminative HMM training with GA for handwritten word recognition," in *Proceedings of the 19th International Conference on Pattern Recognition (ICPR 2008)*, IEEE, Tampa, Fla, USA, December 2008.

[20] B. V. Dhandra, R. G. Benne, and M. Hangarge, "Kannada, Telugu and Devnagari handwritten numeral recognition with probabilistic neural network: a novel approach," *International Journal of Computer Applications*, pp. 83–88, 2010, IJCA special issue on recent trends in image processing and pattern recognition, RTIPPR.

[21] V. J. Dongre and V. H. Mankar, "A review of research on Devnagari character recognition," *International Journal of Computer Applications*, vol. 12, no. 2, pp. 8–15, 2010, (0975-8887).

[22] B. Singh, A. Mittal, and D. Ghosh, "An evaluation of different feature extractors and classifiers for offline handwritten Devnagari character recognition," *Journal of Pattern Recognition Research*, vol. 2, pp. 269–277, 2011.

[23] V. J. Dongre and V. H. Mankar, "Devnagari document segmentation using histogram approach," *International Journal of Computer Science, Engineering and Information Technology*, vol. 1, no. 3, pp. 46–53, 2011.